\documentclass[11pt]{article}

\usepackage{a4wide}
\usepackage{latexsym}
\usepackage{graphicx}
\usepackage{color}
\usepackage[hypertex]{hyperref}

\title{Comparing Typical Opening Move Choices Made by Humans and Chess Engines}

\author{Mark Levene \\
School of Computer Science and Information Systems \\
Birkbeck College, University of London \\
London WC1E 7HX, U.K. \\ mark@dcs.bbk.ac.uk \and Judit Bar-Ilan  \\ Department of
Information Science \\ Bar-Ilan University,  Israel \\
barilaj@mail.biu.ac.il}

\date{}

\begin{document}

\maketitle

\begin{abstract}

The opening book is an important component of a chess engine, and thus computer chess programmers
have been developing automated methods to improve the quality of their books. For chess, which has a
very rich opening theory, large databases of high-quality games can be used as the basis of an
opening book, from which statistics relating to move choices from given positions can be collected.
In order to find out whether the opening books used by modern chess engines in machine versus
machine competitions are ``comparable'' to those used by chess players in human versus human
competitions, we carried out analysis on 26 test positions using statistics from two opening books
one compiled from humans' games and the other from machines' games. Our analysis using several
nonparametric measures, shows that, overall, there is a strong association between humans' and
machines' choices of opening moves when using a book to guide their choices.

\end{abstract}

\section{Introduction}
\label{sec:introduction}

Computer chess has been an active area of research since Shannon's \cite{SHAN50} seminal
paper, where he suggested the basic minimax search strategies and heuristics, which have
been refined and improved over the years. The many advances since then in improving the
search engine algorithms, the static evaluation of chess positions, the representation and
learning of chess knowledge, the use of large opening and endgame databases, and the
exploitation of computer hardware including parallel processing and special-purpose
hardware, have resulted in the development of powerful computer-chess programs, some of
which are of top-Grandmaster-level strength \cite{NEWB02}.

\medskip

The ultimate test for a chess-playing program is to play a match against the World
Champion or one of the leading Grandmasters. In May 1997 an historic match was played
between between IBM's Deep Blue chess computer and Garry Kasparov, then World Champion,
resulting in a spectacular win for the machine, 3.5 -- 2.5. (See \url{www.chess.ibm.com}
for the archived web site of this historic match.) Despite the computer winning the match
there has been an ongoing debate since then on whether the highest ranking computer-chess
programs are at the world-championship level, through most seem to agree that it is
inevitable that eventually chess machines will dominate. Using special-purpose hardware
and parallelization Deep Blue \cite{NEWB02} was capable of analysing up to 200 million
positions per second, while the only other chess program to-date, known to run at a
comparable speed, is Hydra \cite{DONN05a}. The Hydra team, consider their program to be
successor of Deep Blue and their goal is to create the strongest chess-playing computer,
which can convincingly defeat the human world chess champion. A recent six game match
played in June 2005 between Hydra and leading British Grandmaster Michael Adams resulted
in a convincing win for the machine, 5.5 -- 0.5. (See
\url{http://tournament.hydrachess.com} for the web site archiving the match.)

\smallskip

There have been other recent man-machine chess matches against top performing
multiprocessor chess engines capable of analysing several million positions per second,
with the results against world champions still being inconclusive \cite{PRAS03}. It
remains to be seen whether the era of chess man-machine contests is nearing its end,
nonetheless, with machines having ever growing computing resources the future looks bleak
for any human contestants in such matches.

\medskip

Here we concentrate on the late opening/early middle-game phase of the game, and, in
particular the research question we address is whether the opening books used by modern
chess engines in machine versus machine competitions are ``comparable'' to those used by
chess players in human versus human competitions.

\smallskip

For humans, opening preparation is known to be very important, as can be seen, for
example, by the large proportion of chess books concentrating on the opening phase of the
game. Modern chess players also use software packages, such as those developed by
ChessBase (\url{www.chessbase.com}) or Chess Assistant (\url{www.chessassistant.com}), to
assist them in their opening preparation for matches. These packages typically make use of
large databases of opening positions, referred to as {\em opening books}, whose positions
can be searched and are linked to recent databases of games (some of which may be
annotated by experts). This combined with the use of state-of-the-art chess engines for
position analysis, provides players with extremely powerful tools for opening study and
preparation. It is often recommended that chess students combine the study of openings
with typical middle game motifs and endgame structures which may arise from the openings
in question, and computer chess software can be very useful for this purpose
\cite{JACO03}.

\smallskip

Opening theory has become so developed that it is common between expert chess players to
play the first 15 moves or so from memory; see, for example the  Encyclopedia of Chess
Openings marketed by Chess Informant (\url{www.sahovski.co.yu}). As an antidote to the
study of opening theory the former world chess champion, Bobby Fischer, suggested a chess
variant known as Fischer Random chess or Chess960
(\url{www.chessvariants.org/diffsetup.dir/fischer.html}), where the initial position of
the chess pieces is randomised. Due to the 960 different starting positions in Chess960,
knowledge of current chess opening theory is not very useful, and thus the strongest
player will win without having to memorise lengthy opening variations. Chess960 is
becoming a popular variant of chess but, at least in the near future, it is unlikely to
replace classical chess which still fascinates millions world wide.

\medskip

As the top chess engines now compete at Grandmaster level, the opening book has become an
important feature contributing to their success. These days it quite normal for an opening
book specialist to be an integral part of the development team of a chess engine. As an
example of computer chess opening preparation, back in 1995 Fritz 3 defeated a prototype of
Deep Blue in the World Computer Chess Championship when Deep Blue made a crucial mistake as
it went out of its opening book and had to assess the position using its search and
evaluation engine. For its matches against Garry Kasparov in 1997, the Deep Blue team
included Grandmaster Joel Benjamin, who was responsible for developing Deep Blue's knowledge
and fine tuning its opening book \cite{NEWB02}. Since then it has become common practice to
include a Grandmaster level chess player in the chess engine's team especially for high
profile man-machine matches.

\smallskip

Due to the importance of the opening book as a component of a chess engine, computer chess
programmers have been developing automated methods for improving the quality of their books.
For a game such as chess, having a rich and well developed opening theory, a good starting
point for building an opening book is to have available a large database of high-quality
games with a sizeable proportion of recent game records. Game statistics including a summary
of the game results when the move was played, the popularity of the move in the database, how
strong are the players of the move, and how recently the move was played, can be weighted to
produce an evaluation of the ``goodness'' of a move that can inform the chess engine's
evaluation function. Deep Blue utilised these statistics to automatically extend its
relatively small hand-crafted opening book, which consisted of about 4000 positions
\cite{CAMP02}, and, similarly Hydra extends its relatively small opening book, typically
containing about 10 moves per variation (\url{www.hydrachess.com/hydrachess/faq.php}). While
Deep Blue combines the ``goodness'' factors in a non-linear fashion in order to influence the
choice of move in the absence of information from its opening book \cite{CAMP99}, Hydra
combines the ``goodness'' factors in a linear fashion to influence the choice of playable
moves \cite{DONN05b}. An opening book and its extension constructed by the above method will
not be perfect, simply due to the fact that opening theory is still dynamic, and the
statistics often reflect what is fashionable rather than what is objectively best. Hydra
takes this factor into account by adjusting the thinking time of a move when the engine
chooses a book move in preference to a move selected by the engine.

\smallskip

In games such as Awari, where large databases of games do not exist, the above techniques
cannot be applied, so the opening book needs to be constructed by using the game engine to
perform deep searches and generate the evaluation of the positions to be stored in the
book \cite{LINC00}. A best-first search method was proposed by Buro \cite{BURO99}, where
at each level the move with the highest score is expanded first. In this way bad moves are
ignored and no human intervention is necessary. However, this method ignores moves that
are not much worst than the best move, thus allowing an opponent to ``drop out'' of the
book after a few moves and forcing the chess engine to assess positions using its search
and evaluation algorithms. To avoid this situation Lincke \cite{LINC00} proposed to expand
moves in such a way that priority is given to moves at lower search levels, whose score is
within a tolerance level from the best move.

\smallskip

It is also useful to incorporate some form of learning to tune the opening book in order
to avoid playing the same mistake repeatedly. A method developed by Hyatt \cite{HYAT99}
looks at the next ten move evaluations in games it played after the opening book was left,
and extrapolates an approximation of the true value to store in the book as a learnt value
for the position. The learning is conditioned upon the depth of searches that produced the
learnt value, the strength of the opposition in the game played, and whether the engine or
its opponent made a mistake.

\medskip

%The data !!

We now give an overview of the experiment we have carried out. For the purpose of data
analysis we used the {\em Nunn2} test suite devised by Grandmaster Dr. John Nunn to test
chess engines' strength on a variety of late opening/early middle-game positions; the Nunn2
test is distributed by ChessBase together with its chess engine, Fritz. The Nunn2 test was
chosen, since its 25 positions arise from a variety of openings with different
characteristics, and for all these positions there are several reasonable candidate moves. We
augmented the Nunn2 test with the initial position, as we were interested to find out which
first moves do humans and machines prefer.

\smallskip

To compare the choices of humans to those of engines, we made use of two high-quality opening books:
{\em Powerbook 2005}, marketed by ChessBase, derived from a large collection of human versus human
games, and {\em Comp2005} derived from a large collection of machine versus machine games compiled
by  Walter Eigenmann (\url{www.beepworld.de/members38/eigenmann}). For each position we collected
the statistics related to the move choices for the position from both opening books, including the
rank of each move choice, the number of games in the database in which the move choice was played,
and the percentage score achieved for this choice.

\smallskip

%The measures and analysis !!

The analysis we carried out from the data compared both the distribution of the move
choices and the ranks of the choices implied by this distribution. The ranks of the moves
made by humans and engines were compared using a nonparametric association measure we have
used in previous studies, where we compared move choices of different chess engines
\cite{LEVE05} and the ranking of search results by web search engines \cite{BARI05}. The
measure is a weighted version of Spearman's footrule \cite{FAGI03}, which we call the M-measure.
The distributions of move choices were compared using the Jensen-Shannon
divergence (JSD) nonparametric measure \cite{GROSS02,ENDR03}, which allows us to measure
the similarity between two distributions. In addition, for each position we computed the
degree of overlap between move choices of humans and engines and the expected percentage
score for the position, i.e. out of the total number of games played from the position
what was the percentage of wins and draws.

\smallskip

The results show a surprisingly close association between humans' and machines' opening
books. The M-measure is over 0.75 while the JSD is just above 0.70, on average, on a scale
between 0 and 1. It is also shown that, apart from two outliers, the M-measure and JSD are
highly correlated with a correlation coefficient just above 0.65. Moreover, the degree of
overlap between move choices is also just above 0.60 on average, so despite the strong
association between humans' and machines' choice of opening moves there are also
differences, although these disparate moves do not tend to be the highly ranked moves.
Finally, for the positions we investigated, the expected scores from white's point of view
were similar, on average over 55\%, for both humans and machines, which indicates a
significant advantage to white.

\medskip

The rest of the paper is organised as follows. In Section~\ref{sec:measures} we describe
the measures we used to compare the rankings induced by the two opening books. In
Section~\ref{sec:data} we give the detail of the data collection phase. In
Section~\ref{sec:analysis} we present the data analysis carried out and interpret the
results. In Section~\ref{sec:discuss} we discuss possible extensions and applications of
the comparison techniques we have used, and finally, in Section~\ref{sec:concluding} we
give our concluding remarks.

\section{The Measures}
\label{sec:measures}

We used several nonparametric measures \cite{GIBB03} to test the correspondence between
the two opening books.  To illustrate the measures consider the initial chess position and
assume that only the top-10 move choices were recorded in Powerbook 2005 (pb) and in
Comp2005 (comp). (In the experiment 20 moves were actually recorded in each opening book
for the initial position.) The data collected is shown in Table~\ref{table:top10}, where
the second and fourth column indicate the rank of the move in pb and comp, respectively,
while the third and fifth columns indicate the popularity (i.e. the number of games in the
database in which the move was played) in pb and comp, respectively; a zero entry in a
column implies that no games were recorded for that move in the corresponding opening
book.

\begin{table}[ht]
\hskip 12 pt \centering
\begin{tabular}{|c|c|r|c|r|}
\hline  Move & R-pb & P-pb & R-comp & P-comp \\ \hline \hline
e4  & 1 & 448923 & 1 & 122882 \\
d4  & 2 & 361246 & 2 & 105119 \\
Nf3 & 3 & 103542 & 3 &  20820 \\
c4  & 4 &  78408 & 4 &  20023 \\
g3  & 5 &  10142 & 5 &   3359 \\
b3  & 6 &   3252 & 7 &   1121 \\
f4  & 7 &   2754 & 8 &   945  \\
Nc3 & 8 &   1382 & 6 &   1445 \\
b4  & 9 &    718 & 0 &      0 \\
d3 & 10 &    390 & 10 &   333  \\
e3 &  0 &      0 &  9 &   493 \\
\hline
\end{tabular}
\caption{\label{table:top10} Data for top-10.}
\end{table}
\smallskip

The simplest measure we used is the degree of overlap between the two two ranked lists
(Overlap). It can seen from Table~\ref{table:top10} that the overlap is 9 out of 11, i.e.
the degree of overlap is 0.818.

\smallskip

The second measure we used is a weighted variation of Spearman's footrule \cite{FAGI03},
which we call the M-measure \cite{BARI05}. From Table~\ref{table:top10} we see that pb and
comp agree on the ranking of the top-5 move choices but disagree thereafter. To compute
the M-measure we assign each move with a rank greater than zero its reciprocal rank, and
to all moves that did not have a rank, i.e the entry for their rank is zero, we assign a
rank of one plus the maximum rank in the column that had the zero and then take its
reciprocal, as shown in Table~\ref{table:M}.

\begin{table}[ht]
\hskip 12 pt \centering
\begin{tabular}{|c|c|c|}
\hline  Move & R-pb & R-comp \\ \hline \hline
e4  & 1    & 1 \\
d4  & 1/2  & 1/2 \\
Nf3 & 1/3  & 1/3 \\
c4  & 1/4  & 1/4 \\
g3  & 1/5  & 1/5 \\
b3  & 1/6  & 1/7 \\
f4  & 1/7  & 1/8 \\
Nc3 & 1/8  & 1/6 \\
b4  & 1/9  & 1/11 \\
d3  & 1/10 & 1/10 \\
e3  & 1/11 & 1/9 \\
\hline
\end{tabular}
\caption{\label{table:M} Reciprocal ranks for the M-measure.}
\end{table}
\smallskip

To compute the M-measure, we now sum the absolute difference of the the reciprocal ranks
for each pair of corresponding moves as recorded in Table~\ref{table:M}, normalise the
result by dividing by the maximum value of the measure, and finally subtract the result
from one to arrive at a similarity measure rather the a dissimilarity measure. More
formally,  the M-measure, is given by
\begin{displaymath}
1 - \frac{\sum_{i=1}^n \left( \left| \frac{1}{rank_1(i)} - \frac{1}{rank_2(i)} \right|
\right)}{\rm maxM},
\end{displaymath}
where $n$ is the number of moves being compared, $rank_1(i)$ is the rank in pb for the
$i$th move, $rank_2(i)$ is the corresponding rank in comp for the move, and maxM is the
maximum value of the measure to normalise it. The reader can verify that for our
illustrative example, maxM = 4.0398 and M = 0.9694.

\medskip

The third measure we used is the Jensen-Shannon Divergence (JSD) \cite{GROSS02,ENDR03},
which enables us to measure the similarity between two distributions. The first step in
the computation of the JSD is to normalise the number of games played for each move by the
total number of games played. For pb (column 3 of Table~\ref{table:top10}) the
normalisation factor is 1,010,757, and for comp (column 5 of Table~\ref{table:top10}) the
normalisation factor is  276,540. The normalised values, which can be viewed as
probabilities, are shown in Table~\ref{table:JSD}. We denote the pb and comp probabilities
by $p_i$ and $q_i$, respectively; the reader can verify the $p_i$s and the $q_i$s both add
up to one.

\begin{table}[ht]
\hskip 12 pt \centering
\begin{tabular}{|c|c|c|}
\hline  Move & P-pb ($p_i$) & P-comp ($q_i$) \\ \hline \hline
e4  & 0.4441  & 0.4444 \\
d4  & 0.3574 &  0.3801 \\
Nf3 & 0.1024 &  0.0753 \\
c4  & 0.0776 &  0.0724 \\
g3  & 0.0100 &  0.0121 \\
b3  & 0.0032 &  0.0041  \\
f4  & 0.0027 &  0.0034 \\
Nc3 & 0.0014 &  0.0052 \\
b4  & 0.0007 & 0.0000 \\
d3  & 0.0004 & 0.0012 \\
e3  & 0.0000 & 0.0018 \\ \hline
\end{tabular}
\caption{\label{table:JSD} Normalisation of number of games played for the JSD.}
\end{table}
\smallskip

The formal definition of the JSD, which is a symmetric version of the Kullback-Leibler
divergence based on Shannon's entropy, is given by
\begin{displaymath}
1 - \sqrt{\frac{1}{2} \sum_{i=1}^{n_1} p_i \ \log_2{\frac{2 p_i}{p_i + q_i}} + \frac{1}{2}
\sum_{i=1}^{n_2} q_i \ \log_2{\frac{2 q_i}{p_i + q_i}}},
\end{displaymath}
where $n_1$ is the number of moves recorded in pb, $n_2$ is the number of moves recorded
in comp, and the factor of $1/2$ in the square root is in order to normalise the JSD. The
reader can verify that for our illustrative example, JSD = 0.935.

\section{Data Collection}
\label{sec:data}

The positions analysed were the ones from the Nunn2 test collection, augmented with the initial
board configuration as position 26. The move choices made by humans were gathered from Powerbook
2005, which is an opening book marketed by ChessBase, derived from a large collection of high-class
human versus human tournament games. On the other hand, the moves choice made by chess engines were
gathered from Comp2005, which is an opening book we built, derived from a large collection of
high-quality games played between about 2000 chess engines between 2000 and September 2005, with a
time limit of at least 30 minutes per engine per game. The collection was compiled by Walter
Eigenmann and the latest version can be downloaded from his web site,
\url{www.beepworld.de/members38/eigenmann}. (See Table~\ref{table:data} for the detailed numerics
for pb and comp.)

\smallskip

For each opening book and each move choice in a position we analysed, we recorded the rank,
the number of games in the database in which the move was played, and the overall result
achieved when the move was played in terms of the percentage score of wins and draws when
the move was played (a win counts for one point and a draw for half a point).

\begin{table}[ht]
\begin{center}
\begin{tabular}{|l|r|r|} \hline
Opening book & \# Games & \# Positions \\ \hline \hline
Powerbook (pb) &  1,011,611 & 20,070,734 \\
Comp2005 (comp) & 277,288 & 8,239,675 \\ \hline
\end{tabular}
\end{center}
\caption{\label{table:data} Numerics for the opening books}
\end{table}
\medskip

\section{Data Analysis}
\label{sec:analysis}

For each position we computed the M-measure, the JSD and the overlap; the results including
maxM, are shown in Table~\ref{table:summary}. We note that when computing the JSD we have
chosen to discard moves for which less than 10 games were played, as these were deemed to
be statistically insignificant.

\smallskip

The results show a surprisingly close association between humans' and machines' opening
books. The M-measure is over 0.75 while the JSD is just above 0.70, on average, on a scale
between 0 and 1. Moreover, the degree of overlap between move choices is just above 0.60,
on average, so despite the strong association between humans' and machines' choice of
opening moves there are also differences, although the disparate moves do not tend to be
the highly ranked moves.

\smallskip

\begin{figure}[ht]
\centerline{\includegraphics[width=12cm,height=9.33cm]{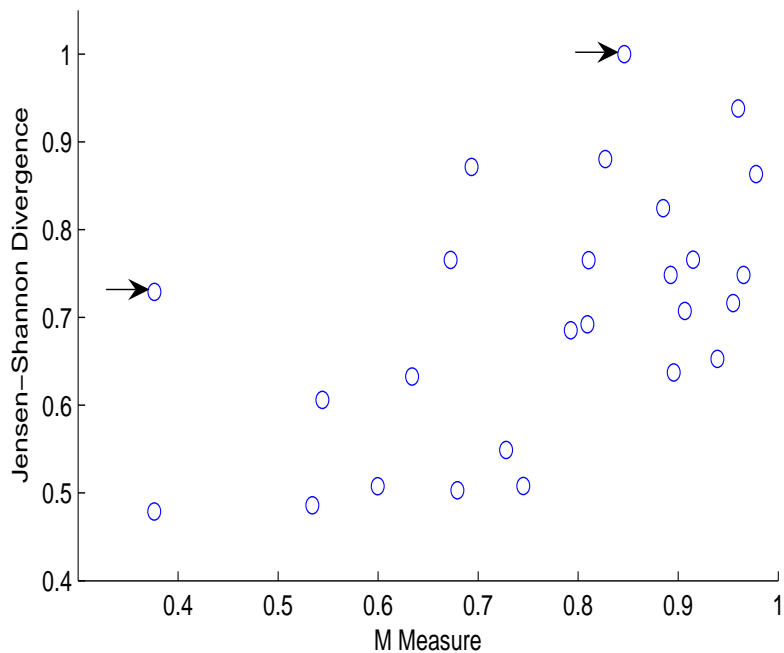}}
\caption{\label{fig:scatter} Scatter plot.}
\end{figure}

The M-measure and the JSD measure are highly correlated as can be seen from their scatter
plot shown in Figure~\ref{fig:scatter}. Their correlation coefficient was computed as
$0.5397$, together with a 95\% bootstrap confidence interval \cite{DAVI97} of $[0.2625,
0.7644]$. Two obvious outliers that stand out in the scatter plot have arrows pointing to
them. The outlier on the left, that has a low M-measure but high JSD, is for position 1
(Symmetrical English opening, Hedgehog variation). In this case, despite the ranking of move
choices being different (low M-measure) their distribution is similar (high JSD). The outlier
on the right is for position 24 (King's Indian Defence, S\"{a}misch variation), for which
there is only one move in each book with more than 9 games played, and thus the JSD is one.
Moreover for this position there were three move choices in pb and only one in comp, so the
M-measure is also very high by default. We measured the correlation coefficient after
removing the outliers and it increased to $0.6549$, together with a 95\% bootstrap confidence
interval of $[0.3655, 0.8228]$.

\smallskip

We observe that Position 15 (Queen's Gambit Declined, Bf4 variation), which is the point
at the bottom left of the scatter plot, is also anomalous in that there is a low
association between pb and comp. The overlap percentage for this position is less that
0.5, which is relatively low. In particular, there are 7 move choices in pb and 12 move
choices in comp, 6 of which overlap. It is interesting to note that this variation is much
more popular with computers than humans, as there were 1022 games recorded for this
positions in comp while only 144 games were present in pb. Moreover, as can be seen in
Table~\ref{table:expected} below, humans perform significantly better in this position
than machines, with an expected percentage score of 57.443\% as opposed to 50.573\%.

\medskip

Table~\ref{table:expected} shows the expected percentage score for each position from the
Powerbook and Comp2005 data sets from white's point of view, and the number of games
recorded in the corresponding databases in which the position occurred. Near each position
number in the first column we indicate whether it was white to move (w) or black (b); it
can be seen that only in 7 out of the 26 positions was it black's turn to move. We note
that, as with the JSD, in our computation of the expected score we have chosen to discard
moves for which less than 10 games were played, as these were deemed to be statistically
insignificant.

\smallskip

For the positions we investigated,  the expected scores from white's point of view,  were
similar, on average over 55\%, for both humans and machines, which indicates a substantial
advantage to white in most of the positions from the Nunn test. Despite this advantage for
white, it is worth noting that the variance of the expected score is rather high; see the last
row in Table~\ref{table:expected}.

\smallskip

It seems that the distribution of move choices is consistent with an exponential
distribution, since there are, generally, only very few popular move choices and the
decrease in popularity is thereafter exponential, although for most positions the number
of choices is too small to reach a definite conclusion. For position 26, for which we have
20 move choices in each book, we fitted an exponential distribution to the sample
distribution induced by the popularity of the move choices in each book, resulting in fits
with $r^2$ (the square of the correlation coefficient) values of 0.9535 and 0.9299,
respectively, for pb and comp.

%In addition, we performed a two-sample Kolmogorov-Smirnov test \cite{GIBB03}, which at the
%1\% level did not reject the hypothesis that the two samples come from the same
%distribution.

\medskip

\begin{table}[ht]
 \hskip 12 pt \centering
\begin{tabular}{|c|c|c|c|c|} \hline
Pos & M-measure & MaxM & JSD & Overlap \\
\hline \hline
1 & 0.376 & 3.085 & 0.729 & 0.714 \\
2 & 0.728 & 3.747 & 0.549 & 0.700 \\
3 & 0.634 & 3.597 & 0.633 & 0.778 \\
4 & 0.745 & 3.091 & 0.508 & 0.500 \\
5 & 0.939 & 3.747 & 0.653 & 0.545 \\
6 & 0.534 & 3.940 & 0.486 & 0.462 \\
7 & 0.915 & 3.019 & 0.766 & 0.571 \\
8 & 0.907 & 3.808 & 0.707 & 0.800 \\
9 & 0.672 & 3.293 & 0.765 & 0.444 \\
10 & 0.693 & 3.112 & 0.871 & 0.556 \\
11 & 0.544 & 3.358 & 0.606 & 0.556 \\
12 & 0.896 & 3.747 & 0.637 & 0.545 \\
13 & 0.827 & 2.700 & 0.880 & 0.500 \\
14 & 0.809 & 3.658 & 0.692 & 0.583 \\
15 & 0.376 & 3.658 & 0.479 & 0.462 \\
16 & 0.811 & 3.635 & 0.765 & 0.700 \\
17 & 0.885 & 3.597 & 0.825 & 0.778\\
18 & 0.955 & 1.850 & 0.716 & 0.500 \\
19 & 0.600 & 3.635 & 0.508 & 0.700 \\
20 & 0.965 & 3.436 & 0.748 & 0.556 \\
21 & 0.978 & 3.019 & 0.863 & 0.833 \\
22 & 0.793 & 3.849 & 0.685 & 0.583 \\
23 & 0.892 & 4.518 & 0.748 & 0.688 \\
24 & 0.846 & 1.083 & 1.000 & 0.333 \\
25 & 0.679 & 3.648 & 0.503 & 0.333 \\
26 & 0.960 & 5.291 & 0.938 & 1.000 \\ \hline
Avg & 0.768 & 3.428 & 0.702 & 0.605 \\
Std & 0.175 & 0.776 & 0.144 & 0.157 \\
\hline
\end{tabular}
\caption{\label{table:summary} Summary of results}
\end{table}

\begin{table}[ht]
\hskip 12 pt \centering
\begin{tabular}{|c|c|r|c|r|} \hline
Pos & Ew\%(pb) & \# pb & Ew\%(comp) & \# comp \\ \hline \hline
1 b & 51.701 & 298 & 60.217 & 138 \\
2 w & 54.473 & 474 & 60.418 & 98 \\
3 w & 52.425 & 414 & 48.529 & 70 \\
4 w & 53.649 & 285 & 62.944 & 674 \\
5 w & 57.968 & 1034 & 58.161 & 1810 \\
6 w & 47.928 & 817 & 52.000 & 111 \\
7 w & 44.000 & 492 & 47.028 & 2810 \\
8 b & 57.616 & 1428 & 59.027 & 297 \\
9 w & 56.468 & 355 & 64.587 & 1002 \\
10 w & 56.361 & 2110 & 56.763 & 1606 \\
11 w & 61.011 & 457 & 68.250 & 72 \\
12 b & 57.578 & 296 & 59.258 & 681 \\
13 w & 50.082 & 743 & 43.656 & 515 \\
14 w & 62.514 & 849 & 60.501 & 439 \\
15 w & 57.443 & 140 & 50.573 & 1011 \\
16 w & 51.387 & 287 & 56.134 & 409 \\
17 b & 54.645 & 307 & 53.581 & 501 \\
18 b & 59.000 & 76 & 60.277 & 303 \\
19 w & 55.243 & 202 & 50.396 & 1272 \\
20 b & 54.180 & 189 & 53.910 & 297 \\
21 b & 55.262 & 420 &53.740 & 2333 \\
22 w & 50.613 & 390 & 51.357 & 143 \\
23 w & 55.029 & 796 & 50.892 & 567 \\
24 w & 60.000 & 63 & 54.000  & 67 \\
25 w & 62.005 & 401 & 60.285 & 397 \\
26 w & 55.036 & 1011597 & 54.749 & 277288 \\ \hline
Avg & 55.139 & $-$ & 55.817 & $-$ \\
Std & 4.301 & $-$ & 5.743 & $-$ \\
 \hline
\end{tabular}
\caption{\label{table:expected} Expected percentage score}
\end{table}

\section{Discussion}
\label{sec:discuss}

Possible extensions and applications of the comparison techniques we have presented are:

\smallskip
\renewcommand{\labelenumi}{(\arabic{enumi})}
\begin{enumerate}

\item Applying the technique to more comprehensive test sets such as the Don Dailey test
\cite{GOMB05}, which consists of 200 positions all of which are 5 moves for each player
from the initial position. A more principled approach could also be taken by collecting
positions from the Encyclopedia of Chess Openings classification system.

\item Comparing opening books of two individuals, be they human or machine. To carry out
such a comparison we need to have game databases of sufficient size, from which we can
construct the respective opening books.

\item Comparing how an opening book changes over time. For example, we could compare
Powerbook 2005 to the newer Powerbook 2006.

\item Extend the technique to middle game and endgame positions with the aid of test sets such
as the WM test of Gurevich and Schumacher of positions from world champion games; the WM
test can be downloaded from \url{www.computerschach.de}. This is more applicable to
comparing the move choices of two available chess engines, which can display the ranking
of the top-n move choices being considered, since, in general, there may be several
reasonable moves from such positions and in game records we have access only to the move that was
chosen.

\item Applying the similarity M-measure to tuning the weights of evaluation function features
such as material balance, mobility, development, pawn structure, and king safety \cite{FURN01} to
those of a specific chess engine. The principle underlying such a technique is to compare, via the
M-measure, the top-n move choices of the evaluation function we are training to the top-n move
choices of the chess engine we are learning from, and to apply a gradient descent (or hill climbing)
method to adjust the weights in the direction of the function we are learning from (cf.
\cite{GOMB05,LUCA06}).

%\item This learning method can also tune evaluation function to that of a human (or for
%that matter to that of a machine) from a database of game records, provided the database
%has a sufficient number of games. In this case we have only a single choice for each move
%in the game record, thus ranked as 1, which we compare to the rank of the same move, this
%time derived from the current state of the evaluation function, say ranked as $m$. In this
%special case the similarity returned by the M-measure would simply be $1/m$.

\end{enumerate}

\section{Concluding Remarks}
\label{sec:concluding}

We have compared the opening books of humans and computers using nonparametric measures.
It seems that there is a strong association between the two books, as the M-measure is
over 0.75 and the JSD just above 0.70, on average. The degree of overlap of move choices is
just above 0.60, on average, so despite the correspondence there are also significant
differences. Moreover, for the positions we investigated, the expected scores from white's
point of view were, on average over 55\%, for both humans and machines, which indicates a
significant advantage to white for the positions we considered.

\smallskip

More experiments need to be carried out on different test sets covering either a wider range
of opening variations or, alternatively, specialising within a small number of popular
opening variations. As mentioned in Section~\ref{sec:discuss} the method we have presented
can be used to compare two individuals' opening choices, be they human or machine. Apart from
a better understanding of the difference between human and machine players such a comparison
may help detect anomalies in an opening book that could be exploited during a match. Finally,
the M-measure, or a refinement of it, does not rely on statistics being readily available, so
could be used as a similarity measure in learning the evaluation function of an opponent.

%\bibliographystyle{alpha}
%\bibliography{game}

\newcommand{\etalchar}[1]{$^{#1}$}

\end{document}